\documentclass[10pt,twocolumn,letterpaper]{article}

\usepackage[pagenumbers]{cvpr} % To force page numbers, e.g. for an arXiv version

\definecolor{cvprblue}{rgb}{0.21,0.49,0.74}
\usepackage[pagebackref,breaklinks,colorlinks,allcolors=cvprblue]{hyperref}
\usepackage{makecell}
\usepackage{multirow}
\usepackage{adjustbox}
\usepackage{colortbl} % 在导言区添加这一行
\definecolor{lightgray}{gray}{0.93} % 定义浅灰色
\def\paperID{****} % *** Enter the Paper ID here
\def\confName{CVPR}
\def\confYear{2026}
\newcommand*\samethanks[1][\value{footnote}]{\footnotemark[#1]}
\title{Omni-Referring Image Segmentation}

\author{
Qiancheng Zheng$^{1}$\thanks{Equal Contribution}, Yunhang Shen$^{2}$\samethanks, Gen Luo$^{3}$, Baiyang Song$^{1}$, Xing Sun$^{2}$,\\
Xiaoshuai Sun$^{1}$, Yiyi Zhou$^{1}$\thanks{Corresponding Author}, Rongrong Ji$^{1}$\\
$^{1}$ Key Laboratory of Multimedia Trusted Perception and Efficient Computing, \\
Ministry of Education of China, Xiamen University, 361005, P.R. China. \\
$^{2}$ Youtu Lab, Tencent, P.R. China.
$^{3}$ OpenGVLab, Shanghai AI Laboratory.\\
{\tt\small \{qianchengzheng,songbaiyang\}@stu.xmu.edu.cn, \{zhouyiyi, xssun, rrji\}@xmu.edu.cn,}
\\
{\tt\small luogen@pjlab.org.cn, shenyunhang01@gmail.com}
}

\begin{document}
\maketitle
\begin{abstract}
In this paper, we propose a novel task termed \emph{Omni-Referring Image Segmentation} (OmniRIS) towards highly generalized image segmentation. Compared with existing unimodally conditioned segmentation tasks, such as RIS and visual RIS, OmniRIS supports the input of text instructions and reference images with masks, boxes or scribbles as omni-prompts. This property makes it can well exploit the intrinsic merits of both text and visual modalities, \emph{i.e.}, \emph{granular attribute referring} and \emph{uncommon object grounding}, respectively. Besides, OmniRIS can also handle various segmentation settings, such as \emph{one v.s. many} and \emph{many v.s. many}, further facilitating its practical use.
To promote the research of OmniRIS, we also rigorously design and construct a large dataset termed \emph{OmniRef}, which consists of 186,939 omni-prompts for 30,956 images, and establish a comprehensive evaluation system. Moreover, a strong and general baseline termed \emph{OmniSegNet} is also proposed to tackle the key challenges of OmniRIS, such as \emph{omni-prompt encoding}.
The extensive experiments not only validate the capability of OmniSegNet in following omni-modal instructions, but also show the superiority of OmniRIS for highly generalized image segmentation. 
The code and dataset will be available at \href{https://github.com/As-Time-Goes-By/OmniSegNet}{link}.
\end{abstract}    
\section{Introduction}
\label{sec:intro}

Recent years have witnessed the great advancement of image segmentation moving from the traditional fixed-category masking to the open-ended ones conditioned on text or visual reference~\cite{jia2022visual,sun2024vrp,suo2024rethinking,li2024visual, jiang2024t}. For instance, the popular \emph{Referring Image Segmentation} (RIS)~\cite{hu2016segmentation}, also known as \emph{Referring Expression Segmentation} (RES)~\cite{luo2020multi}, can segment the referents of arbitrary types in an image according to the given text description, exhibiting much better flexibility than traditional segmentation tasks~\cite{ren2015faster,he2017mask}. Recent RIS tasks, \emph{e.g.}, \emph{Generalized Referring Expression Segmentation} (GRES)~\cite{liu2023gres}, further extend the segmentation patterns to multi-target or no-target grounding. 
More recently, practitioners also resort to visual reference for the identification of similar visual objects in images~\cite{liu2023matcher,li2024visual,zhang2024bridge,xu2025unlocking}, which we term \emph{Visual RIS} here. 

\begin{figure}[t!]
  \centering
  % \fbox{\rule{0pt}{2in} \rule{0.9\linewidth}{0pt}}
    \includegraphics[width=1\linewidth]{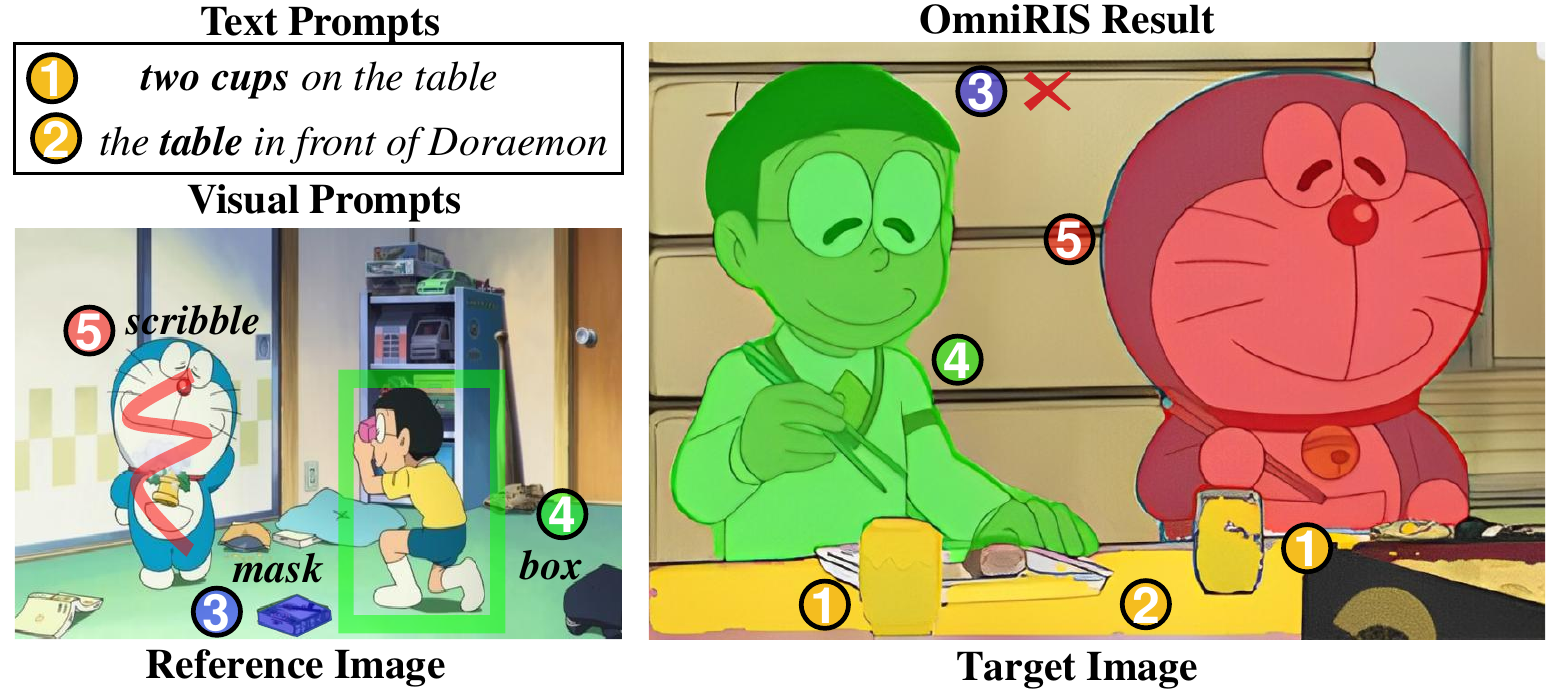}
    \caption{Illustration of the proposed \emph{Omni-Referring Image Segmentation} (OmniRIS) task. OmniRIS can support text expressions (1,2) and reference images with mask (3), box (4) or scribble (5) prompts under the settings of \emph{one v.s. one}, \emph{one v.s. many}, \emph{many v.s. many} or \emph{no-target}.}
   \label{fig1}
   \vspace{-1em}
\end{figure} 

Despite their great successes, existing referring segmentation tasks still have their own shortcomings due to the inherent properties of the unimodal prompts they use. In particular, the text expressions used for RIS can help the model locate common objects with specific attributes or spatial relationships in a rich image. For instance, finding \emph{``two cups on the table''} as shown in \cref{fig1}. However, for the referents that are difficult to describe, RIS models often suffer from low precision~\cite{luo2020multi,yang2022lavt}. In contrast, Visual RIS can well mitigate this issue by referring to a reference image of the target object, also shown in \cref{fig1}. But in order to maintain a high recall rate, existing visual RIS methods~\cite{sun2024vrp,suo2024rethinking} are hard to identify objects of the same category but with different details. In this context, it is natural to question that:
\\
\emph{``Is it plausible to combine the merits of both text and visual references to form a more effective segmentation manner?''}

To approach this target, we propose a novel task termed \emph{Omni-Referring Image Segmentation} (OmniRIS) in this paper, of which illustration is given in \cref{fig1}. As discussed above, OmniRIS can support the inputs of text instructions or reference images, or both, to fully exploit the advantages of the two modalities, \emph{i.e.,} \emph{granular attribute referring} and \emph{uncommon object grounding}, respectively. To correctly identify the referred visual semantics, OmniRIS also requires the model to be aware of masks, boxes or scribbles as additional prompts, which can also facilitate its practical use in terms of human-computer interaction. For a broader scope of application, OmniRIS also supports multiple forms of referring segmentation like GRES~\cite{liu2023gres,xia2024gsva}, \emph{i.e.}, the referring segmentation of \emph{one v.s. many} or \emph{no-target}, while the settings of \emph{many v.s. many} and \emph{many v.s. one} are first introduced to the RIS task. Moreover, OmniRIS also allows the switch between uni- and omni-modal prompts for better generalization. Overall, these properties help OmniRIS to achieve highly generalized image segmentation.

Despite these apparent merits, OmniRIS also has its intrinsic challenges. In addition to multimodal reasoning, a shared problem with previous RIS tasks~\cite{li2024visual,jiang2024t,zou2023segment}, OmniRIS first encounters the issue of how to jointly model the prompts of text and visual modalities. Specifically, a decent OmniRIS model should be capable of learning text expressions and the reference images with spatial prompts at the same time. During inference, the multimodal embedding space it constructs should also allow for the grounding with unimodal or multimodal prompts, posing a key challenge for the multimodal designs of OmniRIS.
Moreover, the training of an OmniRIS model is also intractable due to the flexible but complex task settings. As discussed above, OmniRIS also needs to handle five cases of segmentation which may involve potential referring conflicts, \emph{e.g.}, \emph{many v.s. many}. When given the text and visual prompts simultaneously, the referred objects are often distinct due to the different prompt content. Meanwhile, its other settings, such as \emph{one v.s. one} and \emph{no-target}, also need to be considered. Overall, OmniRIS is a novel but challenging task with great potential for highly interactive segmentation applications.

\begin{figure}[t]
  \centering
    \includegraphics[width=1\linewidth]{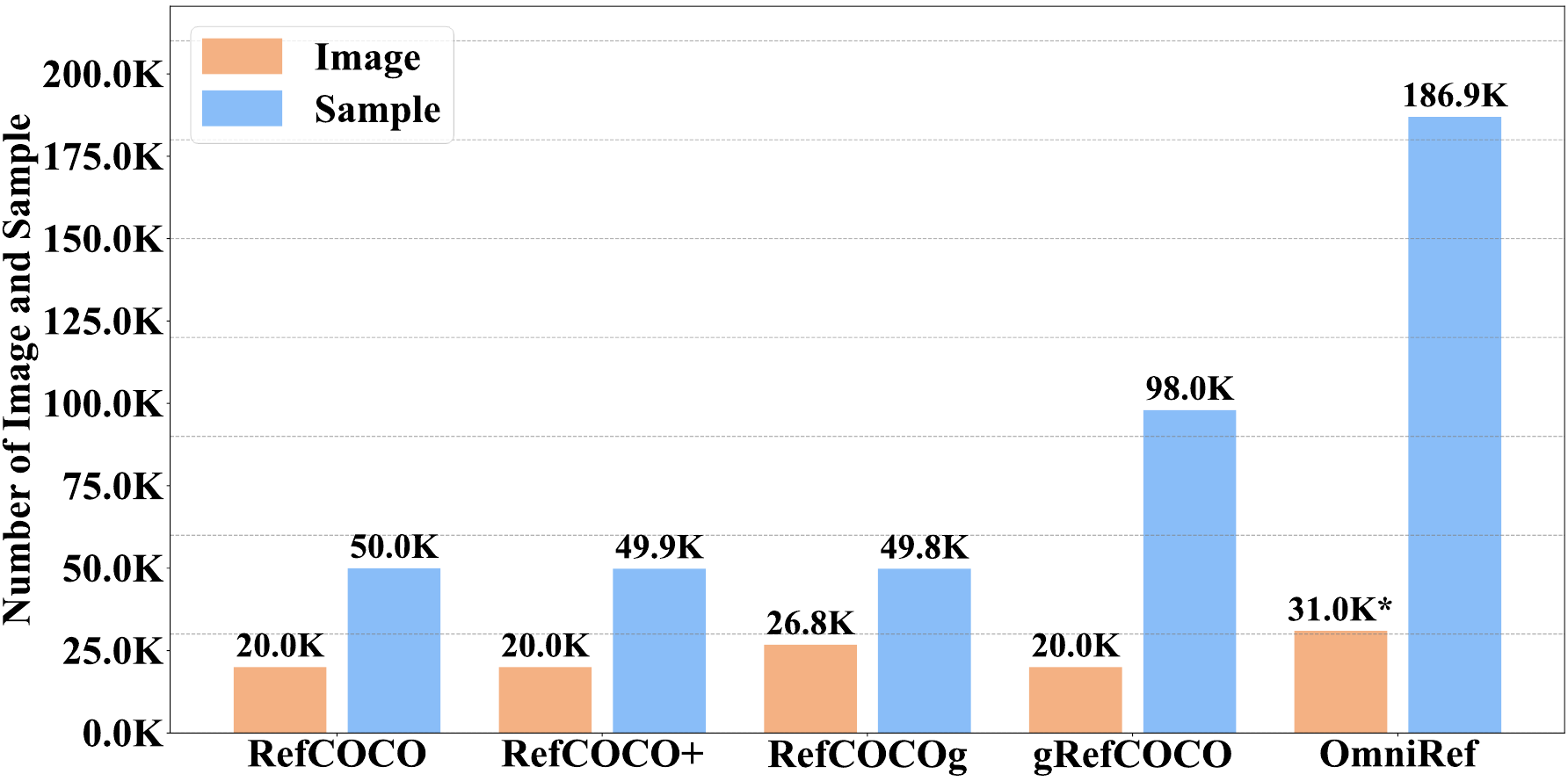}
    \caption{Statistical comparison between our OmniRef dataset and existing popular RIS datasets. *The reference images (26,859) of OmniRef are not counted.}
   \label{fig:ref_compare}
   \vspace{-1em}
\end{figure}

To facilitate the research of OmniRIS, we first propose a large dataset in this paper, termed \emph{OmniRef}. OmniRef consists of $186,939$ omni-prompts for $30,956$ images. Among them, $78,585$ prompts for $6,549$ images are reserved for three test splits, \emph{i.e.}, the \emph{text-only}, \emph{visual-only} and \emph{omni} ones. As shown in \cref{fig:ref_compare}, its scale has surpassed that of existing popular RIS benchmarks, such as RefCOCO/+/g~\cite{yu2016modeling,mao2016generation,kazemzadeh2014referitgame} and gRefCOCO~\cite{liu2023gres}.
% Multi-target samples combine masks of similar objects to handle spatial relationships and occlusions, while single-target samples focus on a single, prominent object for clear recognition. No-target samples provide background support by selecting visually similar reference images without overlapping categories.
% OmniRef consists of four types of samples: \emph{one vs. one}, \emph{one vs. many}, \emph{many vs. many} and \emph{no targets}. It includes referring data with both textual and visual prompts across modalities. As shown in Fig.\ref{dataset_pipeline}, 
The OmniRef dataset is constructed in a rigorous way, which has strict criteria for the selection of images and the annotation of omni-prompts, as shown in \cref{dataset_pipeline}. Based on OmniRef, we also propose a strong baseline model called \emph{OmniSegNet}, which is equipped with a novel \emph{omni-prompt encoder} to handle the issue of omni-prompt modeling and a training regime to tackle the complex settings of OmniRIS.

% Extensive experiments are conducted to validate the strong baseline model OmniSegNet on the proposed OmniRef benchmark. OmniSegNet is also examined on common RIS benchmarks, including  gRefCOCO~\cite{liu2023gres} and RefCOCO/+/g~\cite{mao2016generation,kazemzadeh2014referitgame,yu2016modeling}, confirming its novel designs towards unimodal referring image segmentation. 
Based on the proposed OmniRef dataset, we extensively examine the proposed OmniSegNet on the three test splits, and also compare it with a set of advanced uni-modal RIS methods~\cite{huang2024deep,lai2024lisa,xia2024gsva,liu2023gres}, of which results show its merits in handling unimodal and omni-modal prompts. Moreover, a bunch of qualitative analyses are also conducted to demonstrate the superiority of OmniRIS towards highly generalized image segmentation. 
For instance, the well trained OmniSegNet can be well generalized one-shot segmentation tasks with unseen reference images, as shown in \cref{Qualitative_Analysis}.
% OmniRIS task towards highly-generalized image segmentation.

In summary, our contributions are three-fold:
\begin{itemize}
\item We propose a novel task towards highly generalized image segmentation, termed \emph{Omni-Referring Image Segmentation} (OmniRIS), which supports the inputs of text and visual instructions with omni-prompts.  

\item We construct a large dataset to facilitate the research of OmniRIS, termed \emph{OmniRef}, which contains $186,939$ omni-modal prompts for $30,956$ images.

\item We propose a strong baseline model \emph{OmniSegNet}, which adopts a novel omni-prompt encoder and a carefully designed training regime to overcome the key challenges of OmniRIS.  

\end{itemize}

\section{Related Works}
\subsection{Referring Image Segmentation}
\emph{Referring Image Segmentation} (RIS) denotes the segmentation of target instances in an image according to the given reference information~\cite{yang2024remamber,yu2025latent,cheng2025weakmcn,liu2023polyformer}. In existing literature, RIS mainly refers to the use of text descriptions as the task prompt, \emph{i.e.}, \emph{Referring Expression Segmentation} (RES)~\cite{luo2025cohd,wang2024unveiling,liu2023gres,liu2025hybrid} or \emph{visual grounding}~\cite{xiao2024towards,dai2024simvg,zhu2022seqtr,liu2024grounding,jin2023refclip,sun2023refteacher}. 
Early RIS methods~\cite{hu2016segmentation,luo2020multi} 
often consider RIS as a text-mask matching problem, and adopt a multi-step pipeline with additional relationship feature learning~\cite{shi2018key,li2018referring,yu2018mattnet}. Subsequent research reformulates this task as an end-to-end and cross-modal mask regression problem, and adopts one-stage attention-based models for the joint modeling of multi-modal fusion and reasoning~\cite{liu2023polyformer,wang2022cris,ding2021vision}. 
% For instance, SeqTR considers RIS as a \emph{seq2seq} prediction task, using a lightweight transformer to predict the mask coordinates of referents.
 % Recently, DIT~\cite{huang2024deep} promotes deep interactive learning of text features and image features of the powerful SAM~\cite{kirillov2023segment} through carefully designed deep prompting in the early stages, thus significantly improving the alignment between vision and language.
Recently, a set of works~\cite{xia2024gsva,hu2023beyond,liu2023gres} have been proposed to extend the segmentation settings of RIS. For instance, Liu \emph{et al.}~\cite{liu2023gres} propose the GRES benchmark to extend RIS to more settings, such as \emph{one vs. many} and \emph{no-target} cases. 
\emph{GSVA}~\cite{xia2024gsva} leverages large multimodal models to better handle diverse referring expressions in GRES.
However, these advancements only focus on text-conditioned RIS.

\begin{figure}[t]
  \centering
    \includegraphics[width=1\linewidth]{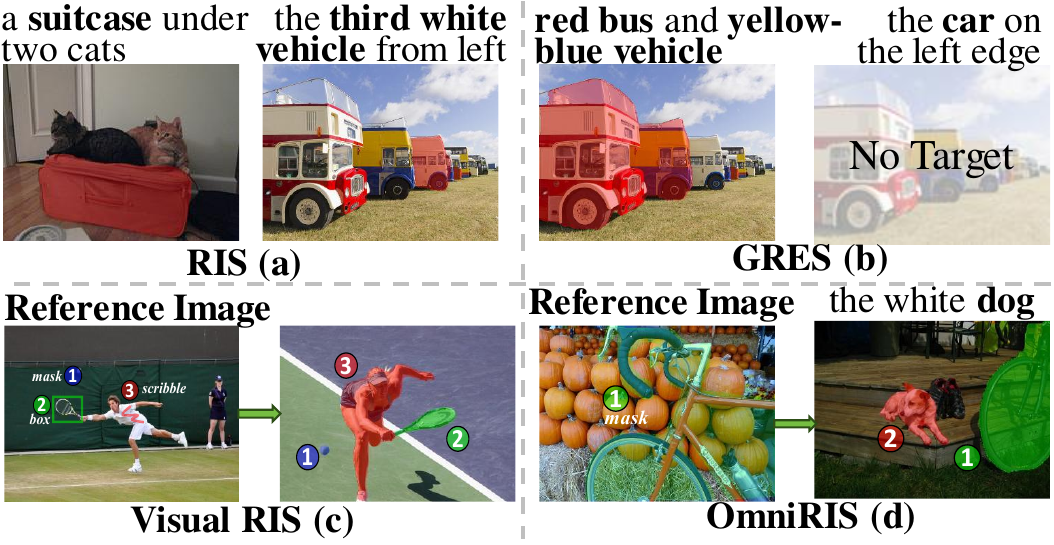}
    \caption{Comparison between OmniRIS and existing unimodal referring segmentation tasks. RIS (a) segments the visual referent corresponding to the given text expression, based on which GRES (b) extends its types of outputs, such as \emph{one v.s. many} and \emph{no-target}. Visual RIS (c) refers to the grounding based on the referenced instances in another image. Compared with existing RIS tasks, the proposed OmniRIS (d) merges the merits of text and visual information to support more flexible image segmentation.}
   \label{fig2}
   \vspace{-1em}
\end{figure} 
\subsection{Visually Referring Image Segmentation}
\emph{Visually Referring Image Segmentation} (Visual RIS) aims to segment objects in a target image based on visual references~\cite{liu2023matcher,suo2024rethinking,zhang2024bridge,hossain2024visual}.
% These references provide explicit visual exemplars and offer a natural modality for representing novel or unseen objects. 
In practice, visual references can be conveyed through various prompt types, such as masks, boxes, or scribbles, and have been widely adopted in one-shot or few-shot segmentation frameworks~\cite{zhang2023personalize,tian2020prior,min2021hypercorrelation,peng2023hierarchical,hong2022cost}. For example, VRP-SAM~\cite{sun2024vrp} enhances SAM~\cite{kirillov2023segment} by introducing visual reference prompts to improve performance in complex scenes. Despite the progress, many existing methods rely on predefined prompt formats and mainly focus on single-image settings, limiting the flexibility and cross-image generalization. Furthermore, mainstream visual prompting techniques~\cite{li2024visual,jiang2024t,zou2023segment,li2024segment} are inferior in capturing consistent cross-image semantics. This shortcoming hinders the alignment and fusion of reference and target features, restricting the effectiveness of Visual RIS in real-world applications.
% Mainstream visual prompt methods~\cite{li2024visual,jiang2024t,zou2023segment,li2024segment} typically use masks, boxes, points or scribbles to indicate regions of interest, often selecting objects within the target image itself as prompts to locate other instances of the same category. Although these methods can indicate the regions of interest to a certain extent, they lack cross-image segmentation capabilities, making it difficult for the model to establish a stable alignment relationship between the target and reference images,  thereby restricting the effective fusion of cross-image features and makes it impossible to complete the Visual RIS task well. 
Compared with text-based RIS, visual RIS is less ambiguous and provides explicit object appearances, but it requires extra reference images or masks and lacks the descriptive flexibility of attribute referring. 
 Overall, omni-modal prompts that aim to unify text and visual prompts have not been effectively handled yet.
% To overcome this, we propose OmniRef, a large-scale textual and visual prompts dataset, and OmniSegNet, a referring segmentation model that incorporates both text and visual prompts to enhance segmentation performance in complex scenarios.
% \input{figs/fig2}
\section{The Definition of OmniRIS}

\begin{figure}[t]
\centering
% \fbox{\rule{0pt}{2in} \rule{.9\linewidth}{0pt}}
\includegraphics[width=1\linewidth]{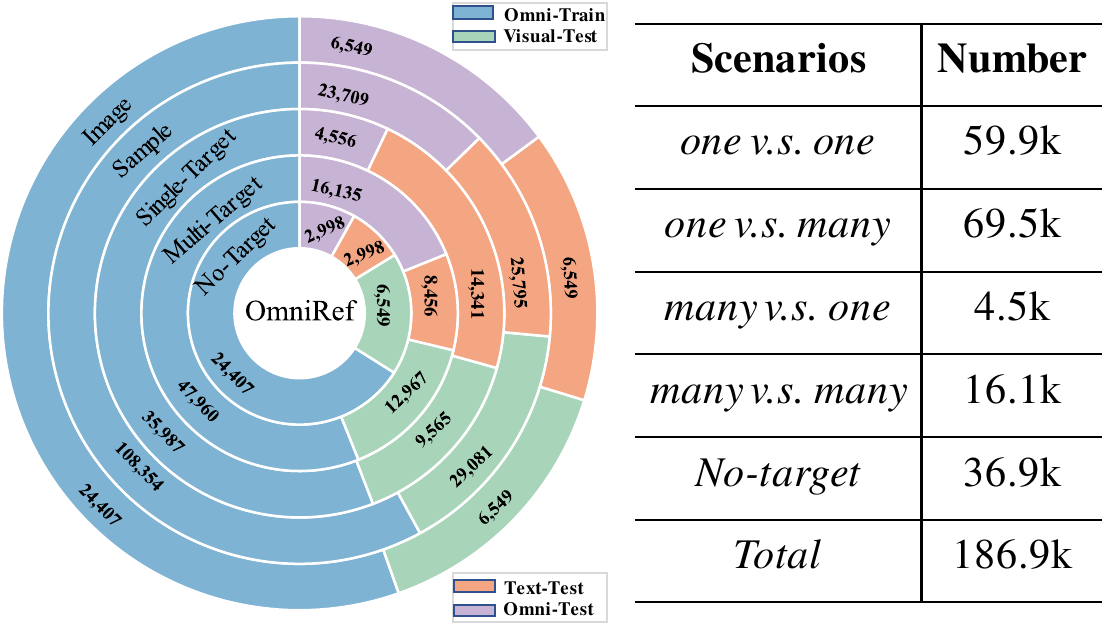}
% \vspace{-3mm}
% \caption{(a) Statistics of the Proposed OmniRef Dataset. (b) Visual Representation of Category Distribution in OmniRef. (c) Examples of Multimodal Referring Prompts in OmniRef.}
\caption{Statistics of the training set and three test splits of OmniRef. Those splits all involve the segmentation outputs of single-target, multi-target and no-target, as well as the cases of \emph{one v.s. one, one v.s. many, many v.s. one}, \emph{many v.s. many} and \emph{no-target}.}
\label{dataset_statistics}
\vspace{-1em}
\end{figure}

\begin{figure*}[t]
% \captionsetup{skip=2pt}
\centering
% \fbox{\rule{0pt}{2in} \rule{.9\linewidth}{0pt}}
\includegraphics[width=1\textwidth]{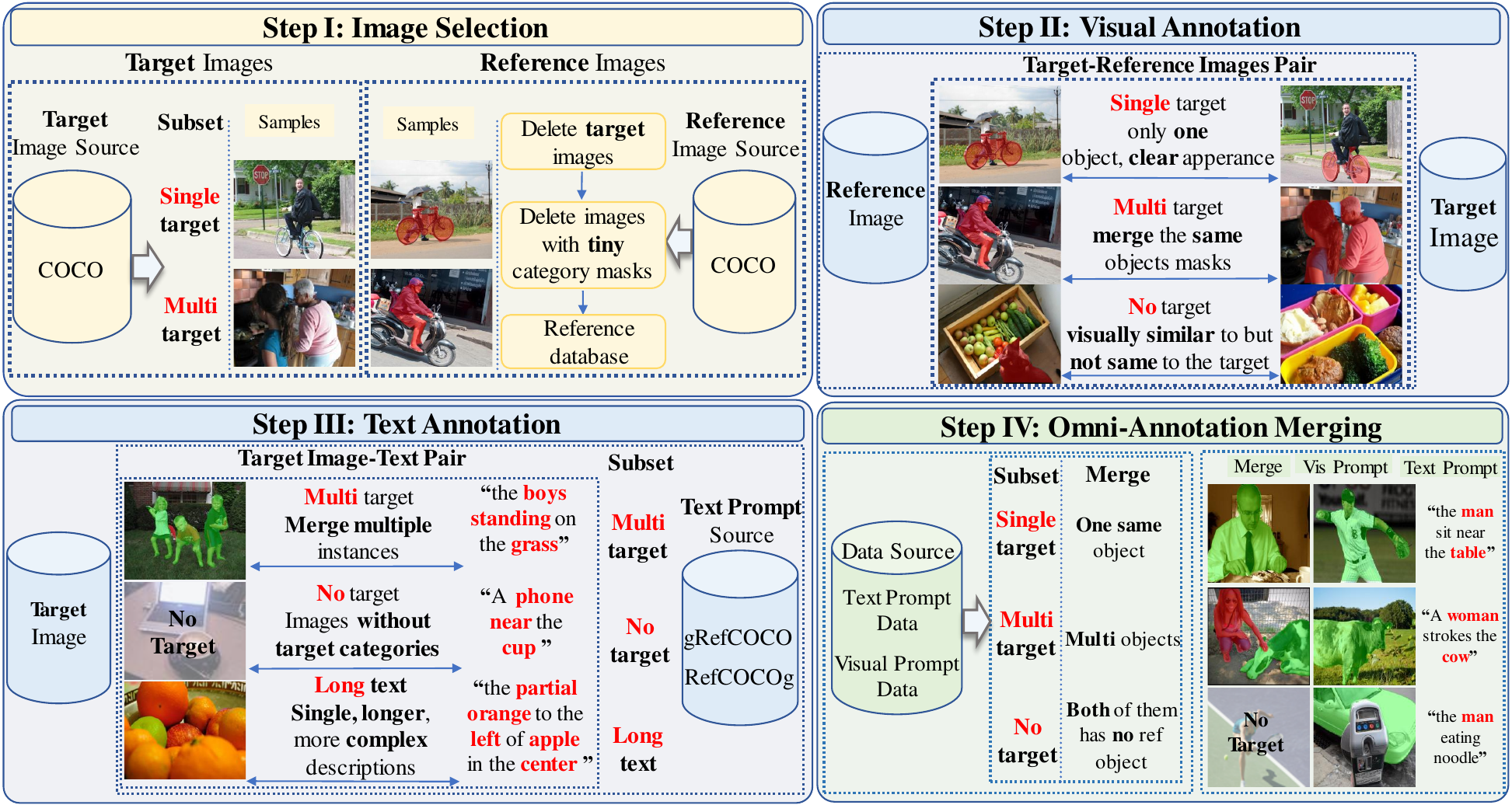}
% \vspace{-3mm}
\caption{The construction pipeline of the proposed OmniRef dataset, which consists of four main steps. Step I filters out the images that have too few objects and lack visual diversity, and selects the ones containing multiple categories and spatially well-distributed objects as the target and reference images. Step II pairs the target and reference images to construct visual prompts for different segmentation cases. Step III aligns text prompts with target images that involve diverse cases, including the \emph{single-target}, \emph{multi-target} and \emph{no-target} outputs with long and complex expressions. Step IV merges visual and text annotations to form the final omni-modal examples. After the four steps, manual checking is also conducted to ensure the quality of examples.}
\label{dataset_pipeline}
\vspace{-1em}
\end{figure*}
In this paper, we propose a novel task for highly generalized image segmentation, termed \emph{omni-referring image segmentation} (OmniRIS). As shown in \cref{fig1} and \cref{fig2}, OmniRIS takes text instructions, reference images or both as input, and it also supports different spatial prompts to specify the referents in images, such as masks, boxes or scribbles.

Concretely, given a target image \( I_t \in \mathbb{R}^{H \times W \times 3} \), OmniRIS supports a set of multimodal prompts $\mathcal{P}$, defined by
\begin{equation}
\mathcal{P} = \{T, (I_r, P_s)\},
\end{equation}
where \( T  \) is a text instruction and \( I_r \in \mathbb{R}^{H \times W \times 3} \) is an reference image. $P_s$ denotes the spatial prompts, all represented as binary matrices $P_s \in \mathbb{R}^{H \times W}$ of the same size as the reference image. These include the segmentation masks $P_{\text{mask}}$, boxes $P_{\text{box}}$ and scribbles $P_{\text{scribble}}$. Note that each element in \( \mathcal{P} \) is optional, enabling the flexible omni-modal interaction.
Thus, the objective of OmniRIS is defined by
\begin{equation}
f_\theta : (I_t, \mathcal{P}) \rightarrow \left( \{M_k\}_{k=1}^K,\, y \right).
\end{equation}
where \( f_\theta \) is a parameterized model. The output consists of $K$ binary segmentation masks \( M_k \in \{0,1\}^{H \times W} \), each corresponding to the objects referred to by the specific prompt in \( \mathcal{P} \), and a binary indicator \( y \in \{0,1\} \) for the existence of the referred objects in the target image. The number of predicted masks \( K \) is adaptively determined based on the number of input prompts.

% where \( f_\theta \) is a parameterized model. The output consists of $K$ binary segmentation masks \( M_k \in \{0,1\}^{H \times W} \), each corresponding to one input prompt in \( \mathcal{P} \). If a prompt refers to multiple objects, its corresponding mask \( M_k \) jointly covers all the referred regions. Therefore, the number of predicted masks \( K \) is determined by the number of input prompts rather than the number of referred objects, and a binary indicator \( y \in \{0,1\} \) denotes the existence of the referred objects in the target image.

From the above definitions, we can see that OmniRIS forms a highly flexible and generalized segmentation manner. In practice, we can use either text or reference images, or both, to mask the referents in the given image. The comparison to existing RIS tasks is illustrated in \cref{fig2}.
\\
\textbf{Main Challenges.} 
Despite the obvious merits in flexibility, OmniRIS is also more challenging than previous RIS tasks. Above all, OmniRIS needs to cope with the prompts of more modalities than previous tasks. And how to help the OmniRIS model be capable of all types of prompts is the first challenge. Moreover, the various segmentation settings also pose more obstacles to the modeling and training of OmniRIS, \emph{e.g.}, enabling the use of single and omni-modal prompts in one model and handling the cases of single-target, multi-target and no-target at the same time. Overall, these properties make OmniRIS a challenging task but one with great potential in various segmentation applications. 

\section{OmniRef: A Large OmniRIS Dataset}

To facilitate the research of OmniRIS, we propose a large dataset in this paper, termed \emph{OmniRef}.
OmniRef has $186,939$ samples from a total of $30,956$ images and the corresponding annotations, such as masks, boxes and scribbles. The detailed statistics of OmniRef are reported in \cref{dataset_statistics}. 
% As shown in Tab.\cref{tab:ref_datasets}, OmniRef significantly surpasses existing RIS benchmarks in scale, enabling broader and more comprehensive evaluation across diverse forms of visual and textual supervision.

\subsection{Dataset Composition}

% The statistics and examples of the proposed OmniRef dataset is reported in Fig.\cref{dataset_statistics}.
% Overall, 
OmniRef has $30,956$ images with $186,939$ omni-prompts. Besides, it also has a high-quality visual reference set consisting of $26,859$ images with $34,570$ instances for visual prompting. These data are divided into the training and three test splits. 
In terms of the training set, \emph{i.e.,} \emph{Omni-Train}, it has $24,407$ images and $108,354$ prompts in total, which includes three types of segmentation outputs, namely \emph{single-target} ($35,987$), \emph{multi-target} ($47,960$) and \emph{no-target} ($24,407$) ones. For the omni-prompts, it has $23,709$ test prompts and $6,549$ reference images, where the latter all include the spatial prompts of masks, boxes and scribbles. 
% Among them, the \emph{many v.s. many} and \emph{no-target} scenarios contain $16{,}135$ and $2{,}998$ samples, respectively.
%
In terms of testing, OmniRIS considers the evaluations of both unimodal and omnimodal RIS. Thus, we build three test splits, namely \emph{Text-test}, \emph{Visual-test} and \emph{Omni-test} splits. These three test sets share a total of $6,549$ images.
The text-test split has $25,795$ text prompts, while the visual-test one has $29,081$ visual prompts. The omni-test split further merges the data of the unimodal ones, including $23,709$ omni-prompts. 
These three test splits are also collected with a balanced distribution of single-target, multi-target and no-target outputs, facilitating the comprehensive evaluation across different application scenarios.

% To build OmniRef, we design a rigorous pipeline to ensure the quality and correctness of its examples. Specifically, xxxxxx #简要介绍构建的过程.
% The construction can be summarized into x stages. 

% \input{figs/dataset_pipeline}

\begin{figure}[t]
\centering
\includegraphics[width=1\linewidth]{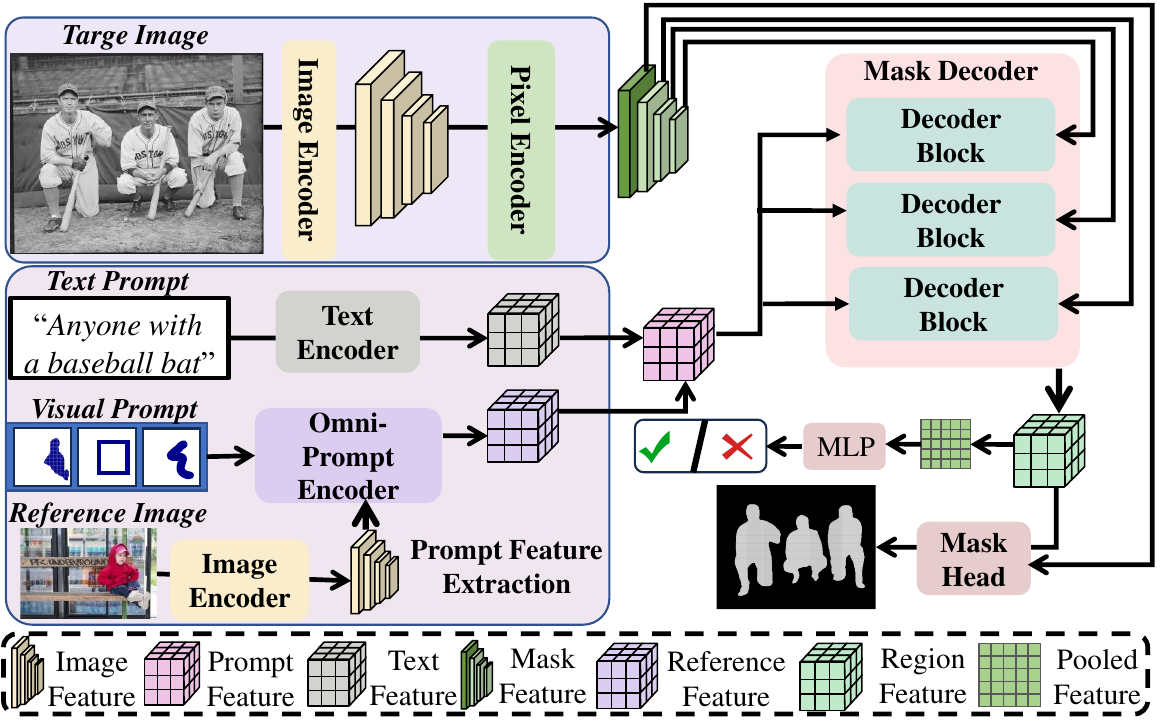}
\caption{The architecture of the proposed OmniSegNet. OmniSegNet introduces a novel omni-prompt encoder to handle the spatial prompts, such as mask, box and scribble. With this omni-prompt encoder, the semantics of the visual referent in the image can be extracted to guide mask decoding. OmniSegNet can also support the processing of visual and text prompts at the same time.}
\label{framework}
\vspace{-1.5em}
\end{figure} 

% ....
\subsection{Dataset Construction}
To build the OmniRef dataset, we design a rigorous four-step pipeline to ensure the quality and correctness of its examples of which process is depicted in \cref{dataset_pipeline}. 
% Specifically, we construct omnimodal referring samples by selecting diverse image scenes, generating high-quality prompts from both text and visual modalities, and associating them with appropriate target objects.
% The OmniRIS dataset construction involves three stages: \emph{Image Selection}, \emph{Visual Annotation}, \emph{Text Annotation} and \emph{Omni-Annotation Merging}.
% The overall process can be categorized into four steps. 
\\
\textbf{Step I: Image Selection.} 
% This stage aims to select target images from existing datasets to construct \emph{OmniRef}. Specifically, target images are selected from the RefCOCO, RefCOCO+ and RefCOCOg datasets, which are built on the COCO dataset and provide rich instance-level annotations. Following the official dataset splits of these referring datasets, OmniRef adopts a consistent partitioning strategy, dividing the data into non-overlapping training and validation sets to ensure fair evaluation and prevent data leakage. 
We first select high-quality images from the popular MSCOCO dataset~\cite{lin2014microsoft}, similar to previous RIS benchmarks~\cite{yu2016modeling, kazemzadeh2014referitgame,liu2023gres}. The selection criterion is that the image content should be rich and diverse enough, thereby providing sufficient semantic context for constructing OmniRIS. To achieve this, we filter out images that contain too few object instances and retain those containing multiple categories and spatially well-distributed objects, ensuring visual complexity. Lastly, we obtain a total of $30,956$ images for the OmniRef dataset.
\\
\textbf{Step II: Visual Annotation.}  
Next, we build the visual reference set with masks, boxes and scribbles as the omni-prompts. Following the previous step, we first collect $26,859$ images from the remaining ones of MSCOCO, and these images all contain obvious visual instances. For these reference images and their visual instances, we then obtain the object masks directly from the instance-level annotations of MSCOCO, ensuring both accuracy and consistency. Based on these masks, we further use the scripts to generate corresponding box and scribble prompts. Specifically, boxes are created by computing the tight bounding rectangles around the mask regions, while scribbles are generated by randomly drawing lines or dots within the mask area. Via this procedure, we create the visual reference set consisting of $34,570$ visual prompt samples.

Lastly, we conduct the annotation of these visual prompts to OmniRef's images. For each target image in OmniRef, we select reference images whose categories match those in the target images, and then form the positive pairs under \emph{multi-target} and \emph{single-target} settings. In the \emph{multi-target} case, the instances of the same category in the target image are merged into one mask, which is then paired with a corresponding reference mask. In the \emph{single-target} case, only one instance is retained. For the \emph{no-target} case, reference images are paired with the target images one by one. 
% These reference images are selected to be visually similar to the corresponding targets but contain object categories that are absent in the target images, thereby serving as challenging negative pairs for robustness evaluation.
These reference images are visually similar to the corresponding targets but contain categories absent from them, serving as challenging negative pairs.
\\
\textbf{Step III: Text Annotation.}  
After the visual annotation, we collect the text prompts for OmniRef's images.
Specifically, we reorganize text prompt samples from the validation and test splits of gRefCOCO~\cite{liu2023gres} and RefCOCOg~\cite{kazemzadeh2014referitgame} based on the target count and instruction complexity. 
Multi-target samples are collected from gRefCOCO by identifying images with text prompts pointing to different object instances. For no-target samples, we extract text prompts whose referred objects are absent in the image.
To enrich the diversity of expression content, we use RefCOCOg to support long-text and single-target evaluation, retaining those samples that refer to a single object using more complex or lengthy descriptions.
Finally, each image in OmniRef is associated with multiple text prompts. 
\\
\textbf{Step IV: Omni-Annotation Merging.}
After the text annotation, we finally merge the omni-modal prompts for OmniRef's images.
We combine the visual and text prompts obtained in the previous two steps that refer to the same target image, thereby constructing omni-modal referring data. Specifically, for \emph{multi-target} cases, the two prompts are required to refer to different object categories within the image. If they refer to the same category, the prompts must jointly cover all its instances. For \emph{single-target} cases, both prompts refer to the same object instance, and the instance must be the only one of its category in the image. For \emph{no-target} cases, the referred objects must not appear in the target image. 
Lastly, each image in the OmniRef evaluation set is paired with both visual and text prompts.

\subsection{Evaluation Protocols}
 In terms of evaluation, OmniRIS includes the metrics of \emph{No-Target Accuracy} (N$_{acc}$)~\cite{liu2023gres}, \emph{Cumulative IoU} (cIoU), \emph{Generalized IoU} (gIoU), and \emph{Pr@X} for comprehensive assessment. N$_{acc}$ measures the performance of no-target samples, cIoU is the \emph{intersection-over-union} ratio, and gIoU averages IoU across all samples. For no-target, true positives/negatives are assigned IoUs of 1/0. Pr@X denotes the percentage of samples whose IoU exceeds a threshold $X$, where no-target samples are excluded. For multi/single-target samples, the threshold is set to $0.7$ due to their larger foreground regions.

\section{The Baseline Model}

\begin{table*}[t]
% \captionsetup{skip=2pt}
\centering
\caption{Comparison between OmniSegNet and existing text and visual RIS methods on three test splits of OmniRef. OmniSegNet exhibits strong and generalized capabilities of both unimodal and omnimodal referring segmentation.}
\small
% \footnotesize
\renewcommand{\arraystretch}{0.9}
% \resizebox{\textwidth}{!}{%
\begin{adjustbox}{width=\textwidth}
\begin{tabular}{l|c|c|ccc|ccc|ccc}
\toprule
\multirow{2}{*}{\makecell{\centering Methods}} &\multirow{2}{*}{\makecell{\centering Backbone}} &\multirow{2}{*}{\makecell[c]{Text\\Encoder}} & \multicolumn{3}{c|}{Text-test} & \multicolumn{3}{c|}{Visual-test}  & \multicolumn{3}{c}{ Omni-test}  \\
\cmidrule(lr){4-6} \cmidrule(lr){7-9} \cmidrule(lr){10-12}
             & &  & cIoU & gIoU & N$_{acc}$ &  cIoU & gIoU & N$_{acc}$  &  cIoU & gIoU & N$_{acc}$ \\
\midrule
\multicolumn{12}{c}{\textit{MLLM Methods}}\\
\midrule            
            \rowcolor{lightgray}
            LISA-7B~\cite{lai2024lisa}& ViT-H & Vicuna-7B & 64.95 & 66.02 & - & - & - & - & - & - & - \\
            \rowcolor{lightgray}
            GSVA-7B~\cite{xia2024gsva} & ViT-H & Vicuna-7B & 65.30 & 67.57 & 63.44 & - & - & - & - & - & - \\

\midrule            
 \multicolumn{12}{c}{\textit{Specialist Methods}}\\
 \midrule
            DIT~\cite{huang2024deep}& ViT-B & BERT & 60.72 & 62.73 & - & - & - & - & - & - & - \\
            ReLA~\cite{liu2023gres} & Swin-B & BERT & 63.40 & 64.75 & 57.97 & - & - & - & - & - & - \\
% \midrule
            GF-SAM~\cite{zhang2024bridge} & ViT-L & - & - & - & - & 40.66 & 42.46 & - & - & - & -\\  
            Matcher~\cite{liu2023matcher} & ViT-L & - & - & - & - & 41.29 & 42.04 & - & - & - & -\\
            VRP-SAM~\cite{sun2024vrp} & ViT-L & - & - & - & - & 55.70 & 52.74 & 78.63 & - & - & -\\
            DCAMA~\cite{shi2022dense} & Swin-B & - & - & - & - & 60.23 & 49.91 & 80.46 & - & - & -\\
            
\midrule
            VRP-SAM+\text{ReLA} & ViT-L & BERT & \underline{63.40} & \underline{64.75} & \underline{57.97} & 55.70 & \underline{52.74} & 78.63 & 61.32 & \underline{59.07} & 55.71\\
            DCAMA+\text{ReLA} & Swin-B & BERT & \underline{63.40} & \underline{64.75} & \underline{57.97} & \underline{60.23} & 49.91 & \underline{80.46} & \underline{62.61} & 58.08 & \underline{55.93}\\
            \textbf{OmniSegNet} & Swin-B & BERT & \textbf{64.92} & \textbf{66.44} & \textbf{62.56} &  \textbf{76.63}& \textbf{68.87} & \textbf{90.81} & \textbf{69.27} & \textbf{67.80} & \textbf{57.69} \\

\bottomrule
\end{tabular}%
\end{adjustbox}

\label{omni-ref-text-visual}
\vspace{-1em}
\end{table*}

In this paper, we also propose a strong and general baseline model for OmniRIS, termed \emph{OmniSegNet}, of which the structure is illustrated in \cref{framework}. 

Concretely, given a target image $I_t$, OmniSegNet first extracts its features $\mathbf{F}_v$ by the vision backbone, and then enhances them via a PixelEncoder: 
\begin{equation}
\mathbf{F}^0_m, \mathbf{F}^i_m = \operatorname{PixelEncoder}(\mathbf{F}_v).
\end{equation} 
$\mathbf{F}^i_m$, where $i \in {1,2,3}$, is used for MaskDecoder, and $\mathbf{F}^0_m$ is for the final mask prediction via MaskHead.
 
In terms of text prompts, text features $\textbf{F}_t$ are extracted by TextEncoder and directly fed to MaskDecoder. 
For reference images, they are first processed by the vision backbone to obtain the visual features $\mathbf{F}_r$.
Afterwards, the visual features and the different spatial prompts are jointly modeled by the proposed \emph{Omni-Prompt Encoder} consisting of a \emph{Prompt Embed Module} (PEM) and the \emph{Prompt Generator}:
% \begin{equation}
% \mathbf{F}_{vs} = \operatorname{PEM}(\mathbf{F}_r, \mathbf{P}),
% \end{equation}
% \begin{equation}
% \mathbf{F}_{p} = \operatorname{PromptGenerator}(\mathbf{F}_{vs}, \mathbf{F}_q).
% \end{equation}
\begin{equation}
\mathbf{F}_{p} = \operatorname{PromptGenerator}\big(\operatorname{PEM}(\mathbf{F}_r, \mathbf{P}), \mathbf{F}_q\big).
\end{equation}
% \begin{align}
% \mathbf{F}_{vs} &= \text{PEM}(\mathbf{F}_r, \mathbf{P}_s), \\
% \mathbf{F}_{p} &= \text{PromptGenerator}(\mathbf{F}_{vs},\mathbf{F}_q).
% \end{align}
Here, $\mathbf{P}$ denotes the spatial prompt from the reference image, $\mathbf{F}_{vs}$ is the fused visual reference features, $\mathbf{F}_q$ are learnable queries and $\mathbf{F}_p$ is the final prompt embedding used to guide segmentation.

Lastly, $\mathbf{F}_t$ and $\mathbf{F}_p$ are combined and fed to MaskDecoder, as shown in \cref{framework}. Due to the page limit, the details of our OmniSegNet are given in our \textbf{Appendix}.
\\
\textbf{Training Regimes.}
To tackle the complex settings of OmniRIS, we further introduce a three-step training regime for our OmniSegNet. \emph{i. Vision-Language Alignment.} We first conduct the VL pretraining using the text RIS dataset, which aims to help OmniSegNet be capable of both instruction understanding and conditional segmentation. \emph{ii. Visual Instruction Tuning.} Afterwards, we fix the text encoder and learn to project the visual prompt features onto the semantics of OmniSegNet using the visual prompt data of OmniRef. Via this process, OmniSegNet learns to be aware of visual conditions for segmentation. \emph{iii. Joint Omni-modal Training.} After the second step, OmniSegNet has been capable of both visual and text referring, but with degradations in text referring. Thus, we mix the omni-data for the joint training of OmniSegNet. The details can refer to our code project given in the supplementary materials.
\section{Experiments}

\begin{table}[t]
% \captionsetup{skip=2pt}
\renewcommand{\arraystretch}{0.85}
    \centering
    \caption{Ablation study of OmniSegNet on the Visual-test of the OmniRef. (a) shows the results of using different visual prompts. (b) ablates the settings of the prompt embed module.}
    \resizebox{\columnwidth}{!}{ % 让表格适应单栏

    \begin{tabular}{c|ccccc}  
        \toprule  
        Configuration & cIoU & gIoU & Pr@0.7 & Pr@0.8 & Pr@0.9 \\  
        \midrule  
        \multicolumn{6}{c}{(a) Types of Visual Prompts} \\  
        \midrule  
        \emph{scribble} & 73.20 & 63.31 & 47.17 & 35.85 & 15.08 \\  
        \emph{box} & 74.00 & 64.21 & 48.84 & 36.71 & 14.99 \\  
        \emph{mask} & \textbf{75.35} & \textbf{66.36} & \textbf{52.68} & \textbf{40.61} & \textbf{18.30} \\  
        \midrule
        \multicolumn{6}{c}{(b) Prompt Embed Module Fusion Methods} \\  
        \midrule  
        \emph{-} & 62.61 & 55.63 & 40.36 & 30.10 & 11.77 \\  
        \emph{multiply} & 74.61 & 64.31 & 49.39 & 37.52 & 16.27 \\  
        \emph{add} & \textbf{75.35} & \textbf{66.36} & \textbf{52.68} & \textbf{40.61} & \textbf{18.30} \\  
        \bottomrule  
    \end{tabular} 

    }

    \label{ablation_omni_ref}
    \vspace{-1em}
\end{table}

\begin{table*}[t]
\centering

\caption{Evaluation of OmniSegNet on common RIS benchmarks. OmniSegNet achieves competitive performance under the Text-RIS evaluations, showing general performance for Text-RIS tasks.}
\small
% \footnotesize
\renewcommand{\arraystretch}{0.85}
\resizebox{\textwidth}{!}{%

\begin{tabular}{l|c|ccc|ccc|ccc|cc}
\toprule
\multirow{2}{*}{\makecell{\centering Methods}} & \multirow{2}{*}{\makecell{\centering Backbone}} & \multicolumn{3}{c|}{gRefCOCO}& \multicolumn{3}{c|}{RefCOCO} & \multicolumn{3}{c|}{RefCOCO+} & \multicolumn{2}{c}{RefCOCOg}   \\
\cmidrule(lr){3-5} \cmidrule(lr){6-8}\cmidrule(lr){9-11} \cmidrule(lr){12-13}
                &  &      val & testA & testB &      val & testA & testB & val & testA & testB & val-u & test-u  \\
\midrule

 \multicolumn{13}{c}{\textit{MLLM Methods}}\\
 \midrule
            \rowcolor{lightgray}
            PixelLM-7B~\cite{ren2024pixellm} & CLIP-ViT-L  & -&-&-&73.00 & 76.50 & 68.20 & 66.30 & 71.70 & 58.30 & 69.30 & 70.50 \\
            \rowcolor{lightgray}
            LISA-7B~\cite{lai2024lisa} & ViT-H & 61.76&68.50&60.63& 74.90 & 79.10 & 72.30 & 65.10 & 70.80 & 58.10 & 67.90 & 70.60\\
            \rowcolor{lightgray}
            GSVA-7B~\cite{xia2024gsva} &ViT-H&63.29&69.93&60.47&77.20&78.90&73.50&65.90&69.60&59.80&72.70&73.30\\
\midrule

 \multicolumn{13}{c}{\textit{Specialist Methods}}\\
 \midrule
            DIT~\cite{huang2024deep} & ViT-B  &-&-&-& 71.98 & 74.51& 68.77 & 59.97 & 65.52 & 51.72 & 60.18 & 61.15 \\
% \midrule
            LAVT~\cite{yang2022lavt} & Swin-B   &57.64&65.32&55.04& 72.73 & 75.82 & 68.79 & 62.14 & 68.38 & 55.10 & 61.24 & 62.09  \\
% \midrule
            ReLA~\cite{liu2023gres} & Swin-B &62.42&\underline{69.26}&\underline{59.88} & 73.82 & 76.48 & 70.18 & \textbf{66.04} & 71.02 & \underline{57.65} & \textbf{65.00} & 65.97  \\
            % DMMI~\cite{hu2023beyond} & Swin-B & 74.13 & 77.13 & 70.16 & 63.98 & 69.73 & 57.03 & 63.46 & 64.19  \\
% \midrule
            LQMFormer~\cite{shah2024lqmformer} & Swin-B  & \underline{64.98} &-&-& \textbf{74.16} &  \underline{76.82} & \textbf{71.04} & 65.91 & \textbf{71.84} & 57.59 & 64.73 & \underline{66.04}  \\
\midrule
            OmniSegNet & Swin-B &\textbf{65.30}&\textbf{70.98}&\textbf{62.18}& \underline{74.02} & \textbf{76.93} & \underline{70.63} & \underline{65.95} & \underline{71.30} & \textbf{58.31} & \underline{64.90} & \textbf{66.27}  \\
            % OmniSegNet^{\ast} & Swin-B&\textbf{65.30}&70.98&
            % \textbf{62.18}&-&-&-&-&-&-&-&-\\

\bottomrule
\end{tabular}%

}

\label{grefcoco-refcoco}
\vspace{-1em}
\end{table*}
\begin{table}[t]
\centering
\caption{One-shot Semantic Segmentation results (mIoU) on PASCAL-5$^i$ and LVIS-92$^i$ datasets.}
\resizebox{\columnwidth}{!}{%
\begin{tabular}{l|ccccc|c}
\toprule
\multirow{2}{*}{Methods} & \multicolumn{5}{c|}{PASCAL-5$^i$} & \multicolumn{1}{c}{LVIS-92$^i$} \\
\cmidrule(lr){2-6} \cmidrule(lr){7-7}
 & 5$^0$ & 5$^1$ & 5$^2$ & 5$^3$ & Mean & Mean \\
\midrule
PerSAM~\cite{zhang2023personalize} & - & - & - & - & 43.1 & 11.5 \\
PerSAM-F~\cite{zhang2023personalize} & - & - & - & - & 48.5 & \textbf{12.3} \\
Painter~\cite{wang2023images} & - & - & - & - & - & 10.5 \\
MIAPnet~\cite{pandey2023weakly} & 45.9 & 46.9 & 47.2 & 41.5 & 45.4 & - \\
\midrule
OmniSegNet & \textbf{56.0} & \textbf{52.9} & \textbf{53.8} & \textbf{49.3} & \textbf{53.0} & 9.2 \\
\bottomrule
\end{tabular}
}
\label{tab::one-shot}
\vspace{-1em}
\end{table}

% We further examine the proposed OmniSegNet on the proposed OmniRef benchmarks, and also compare it with a set of advanced RIS models on the three test splits.  

\subsection{Quantitative Analysis}

We first evaluate the set of visual and textual RIS methods as well as our OmniSegNet on the proposed OmniRef benchmark.
% In practice, we only report these unimodal RIS methods on the sub-benchmark of OmniRef due to their settings, and report the full results of OmniSegNet on all splits.
Since existing RIS methods can only perform unimodal RIS tests, we also introduce two strong OmniRIS baselines in \cref{omni-ref-text-visual}, \emph{i.e.}, ReLA+VRP-SAM and ReLA+DCAMA. In these two baselines, ReLA is in charge of text examples of OmniRef, while VRP-SAM and DCAMA are used to process visual RIS examples.
In terms of the text prompt evaluations, \emph{i.e.}, \emph{Text-test}, the MLLM-based method GSVA-7B~\cite{xia2024gsva} achieves the best performance on this test split, but it adopts much larger backbones than the other methods, \emph{i.e.}, ViT-H~\cite{dosovitskiy2020image} and the 7B-level LLM called Vicuna~\cite{chiang2023vicuna}. 
% In contrast, OmniSegNet adopts the more lightweight Swin-B~\cite{liu2021swin} backbone and BERT~\cite{devlin2018bert} language model, highlighting its superior efficiency. 
% Compared with ReLA, which uses the same backbone and language model as OmniSegNet, our method consistently improves the performance on all evaluation metrics, demonstrating its effectiveness in text-based referring image segmentation.
For the methods using the same-scale backbones, we can see that OmniSegNet can achieve better performance on this evaluation, showing its great capability in following text instructions.
In the second block of \cref{omni-ref-text-visual}, we compare OmniSegNet with four advanced visual RIS methods on the visual-test split. Since these methods cannot identify the existence of objects, we add a classification head to VRP-SAM and DCAMA and retrain them on OmniRef. Compared with these advanced visual RIS methods, our OmniSegNet also shows better performance on all three metrics, indicating its superiority in handling visual prompts.
%
% For the omni-test, we combine ReLA’s text-based predictions with the visual predictions from VRP-SAM or DCAMA to obtain the final baseline results.
% Compared with these combined baselines, OmniSegNet achieves significantly better performance on the omni-prompt setting, demonstrating its ability to seamlessly integrate and reason over both modalities.
For the Omni-test, OmniSegNet surpasses the combined baselines ReLA+VRP-SAM and ReLA+DCAMA by a large margin, demonstrating its strong capability to seamlessly model and reason over both textual and visual modalities.
Overall, these demonstrate the effectiveness and generalization of OmniSegNet.

%
%  In the last block, we also report the performance of OmniSegNet on the omni-prompt setting, which achieves a performance of \emph{70.93} cIoU / \emph{69.34} gIoU / \emph{57.88} N$_{acc}$ on this task, verifying its versatility and omnimodal ability. 
% Overall, these demonstrate the effectiveness and generalization of OmniSegNet.

\subsubsection{Ablation Study of OmniSegNet}
% \paragraph{Ablation Study of OmniSegNet.}
% \input{tables/one-shot}
% \input{tables/ablation}
% \input{tables/ratio}
% \input{tables/grefcoco-refcoco}
% \input{tables/one-shot}

\begin{figure*}[t] 
\centering
% \fbox{\rule{0pt}{2in} \rule{.9\linewidth}{0pt}}
\includegraphics[width=1\textwidth]{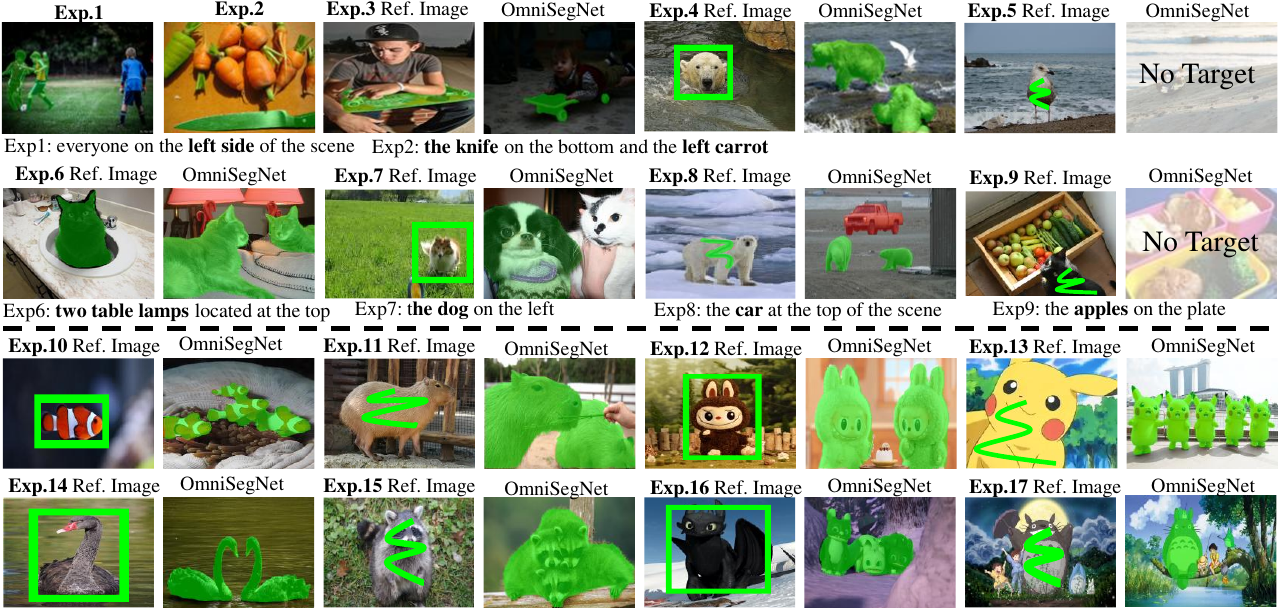}

\caption{Visualized predictions of OmniSegNet. The first block (Exp.1-9) shows text-only, visual-only and omnimodal samples from the OmniRef dataset, reflecting diverse scenarios such as \emph{one v.s. one} (Exp.2), \emph{one v.s. many} (Exp.1-2,4), \emph{many v.s. one} (Exp.7), \emph{many v.s. many} (Exp.6,8) and  \emph{no-target} (Exp.5,9). The last block (Exp.10–17) demonstrates the one-shot capability of OmniSegNet in handling unseen visual instances.}

\label{Qualitative_Analysis}

% \vspace{-0.5em}
\end{figure*}

% To understand the key components and design choices of OmniSegNet, we conduct ablation studies on three critical aspects, \emph{i.e.}, the types of visual prompts, the design of the \emph{Prompt Embedding Module} (PEM), and the sampling ratio between text- and visual-guided data during training.
To understand the key components and design choices of OmniSegNet, we conduct ablation studies on three aspects: the types of visual prompts, the design of the \emph{Prompt Embedding Module} (PEM) and the sampling ratio between text- and visual-guided data during training, whose detailed results are presented in the \textbf{Appendix}.
\\
In \cref{ablation_omni_ref} (a), we ablate the results of using different visual prompts.
% The experimental results clearly show that the performance of \emph{mask}, \emph{box} and \emph{scribble} gradually decreases.
Specifically, \emph{mask} provides the most comprehensive information about object features and effectively preserves the context of the reference object, so it performs best among all types of prompts.
% In contrast, since \emph{box} contains background information, some noise affects the learning process of object features, resulting in performance degradation.
In contrast, \emph{box} presents the suboptimal performance, mainly due to the noises contained in its box backgrounds.
Similarly, \emph{scribble} cannot fully capture the features of the object due to its irregular shape and limited coverage, resulting in the lowest performance.
These observations indicate that the quality and completeness of visual prompts are important for OmniRIS.
In \cref{ablation_omni_ref} (b), we examine different fusion designs of the \emph{Prompt Embedding Module} (PEM). Three settings are compared, which are \emph{no fusion} (-), \emph{multiplication-based fusion} (multiply) and \emph{addition-based fusion} (add).
When no fusion is applied, \emph{i.e.}, the PEM is absent, there is no interaction between the reference visual features and the prompt embedding features. As a result, the model fails to accurately capture the characteristics of the target object. This implies the importance of the  PEM in guiding the visual referring segmentation process.
Furthermore, the method used to fuse prompt and reference features also affects performance. Compared to multiplication, the addition operation yields better results across all metrics. This can be attributed to the fact that addition provides a more stable and flexible feature integration, while multiplication may lead to information suppression.
% \textbf{Effect of Batch Sample Ratio.}
% \cref{ratio} presents a comparative analysis of using different batch sample ratios during joint training of the gRefCOCO dataset and visual prompt data from the OmniRef benchmark. The evaluation datasets are their respective test sets, and the impact of varying ratios on gRefCOCO-val and Visual-test performance is assessed.
% %
% For gRefCOCO-val, increasing the proportion of text-guided samples significantly improves performance, indicating that rich language supervision helps the model better understand and express natural language. In Visual-test, while adding visual-guided samples is helpful, performance changes are relatively stable, with slight differences between proportions.
% %
% In summary, the \emph{7/2} ratio achieves the best performance in both tasks, balancing text and visual prompt learning, enhancing the model's performance, and confirming the value of strategic data distribution in multimodal segmentation training。

\subsubsection{Generalization of OmniSegNet}
% \paragraph{Generalization of OmniSegNet.}

To validate the generalization of the proposed OmniSegNet, we also evaluate it on existing RIS benchmarks in \cref{grefcoco-refcoco}.
% To validate OmniSegNet's generalization, we evaluate it on existing RIS benchmarks.
As shown in \cref{grefcoco-refcoco}, on gRefCOCO, OmniSegNet surpasses the previous SOTA method ReLA with absolute gains of 2.88\%, 1.72\% and 2.30\% on the three evaluation sets, respectively. Notably, it even outperforms the powerful MLLM-based methods such as LISA-7B~\cite{lai2024lisa} and GSVA-7B~\cite{xia2024gsva} with absolute improvements of 2.01\%, 1.05\% and 1.55\% on the splits of gRefCOCO, respectively. 
In addition, OmniSegNet performs well on RefCOCO/+/g, providing results comparable to the SOTA methods. On RefCOCO and RefCOCO+, the performance of our model is comparable to the MLLM-based methods such as LISA-7B~\cite{lai2024lisa} and PixelLM-7B~\cite{ren2024pixellm}.
Experimental results show that OmniSegNet performs strongly in text referring segmentation, achieving state-of-the-art results among specialist methods and delivering comparable or even superior accuracy to MLLM-based approaches.
% \textblue{Experimental results show that OmniSegNet performs strongly in text referring segmentation, with its lightweight configuration achieving similar or better accuracy than the MLLM method.}
%

To further examine the visual RIS capability, we also test OmniSegNet on One-shot Semantic Segmentation~\cite{pandey2023weakly,wang2023images,zhang2023personalize}, even though this is not its primary design focus. To evaluate its potential in this task, we conduct experiments on the PASCAL-5$^i$~\cite{everingham2010pascal} and LVIS-92$^i$~\cite{gupta2019lvis} datasets. As shown in \cref{tab::one-shot}, OmniSegNet achieves competitive performance, particularly on PASCAL-5$^i$, where it outperforms other comparison methods. Overall, these results well validate the generalization and designs of OmniSegNet.

\subsection{Qualitative Analysis}

To gain insights into the proposed OmniSegNet and OmniRIS task, we also visualize a set of predictions in \cref{Qualitative_Analysis}. The experiments (Exp.1–17) reflect OmniSegNet’s capabilities across the diverse scenarios of the OmniRIS task, including \emph{one v.s. one}, \emph{one v.s. many}, \emph{many v.s. one}, \emph{many v.s. many} and  \emph{no-target} cases.
For the text-only examples (Exp.1–2), the model exhibits strong spatial reasoning and accurately grounds the referents based on complex instructions, such as ``\emph{everyone on the left side of the scene}'' or ``\emph{the knife on the bottom and the left carrot}'', demonstrating its effectiveness in \emph{one v.s. many} text-based segmentation.
In the visual-only cases (Exp.3–5), OmniSegNet responds precisely to different types of visual prompts, including masks, boxes and scribbles. For example, given a reference image with a \emph{box} indicating a \emph{bear} (Exp.4), the model precisely segments the corresponding targets in the target image, highlighting its capability in \emph{one v.s. one}, \emph{one v.s. many} and  \emph{no-target} visual scenarios.
Omnimodal examples (Exp.6–9) highlight OmniSegNet’s unique ability in processing omnimodal prompts by jointly leveraging textual and visual information, including \emph{many v.s. one}, \emph{many v.s. many} and  \emph{no-target} scenarios. Given the expression along with a corresponding reference image, the model produces semantically aligned and spatially accurate segmentation, showcasing its omnimodal fusion capability. 
Lastly, the examples in the last block (Exp.10-17) reflect OmniSegNet’s strong one-shot ability. When given unseen objects during testing, the model can still reliably locate and segment targets using only the provided prompts. 
% These qualitative results confirm OmniSegNet’s comprehensive capabilities in the OmniRis task and one-shot referring segmentation scenarios. Moreover, these examples also show the merits of OmniRIS towards highly generalized segmentation.
These qualitative results confirm OmniSegNet’s comprehensive capabilities in the OmniRIS task and one-shot referring segmentation scenarios. Moreover, these examples highlight the merits of OmniRIS for highly generalized segmentation.
\section{Conclusion}
In this paper, we propose a novel task termed OmniRIS. Compared with existing unimodal RIS tasks, OmniRIS aims to merge the merits of both text and visual modalities to achieve highly generalized segmentation. In practice, OmniRIS can support the input of text descriptions and reference images with omni-prompts, such as masks, boxes and scribbles. Moreover, OmniRIS includes various segmentation settings, such as \emph{many vs. many}. To support the research of OmniRIS, we also build a large dataset called OmniRef, based on which a strong and general baseline model called \emph{OmniSegNet} is also proposed in this paper. Extensive experiments and analyses not only validate the advantages of OmniSegNet in handling both text and visual RIS tasks but also confirm the merits of OmniRIS towards highly generalized image segmentation.
{
    \small
    \bibliographystyle{ieeenat_fullname}
    \bibliography{main}
}

% WARNING: do not forget to delete the supplementary pages from your submission 
% \appendix
\clearpage
% \setcounter{page}{1}
% \maketitlesupplementary
\appendix

\begin{figure*}[t]
\centering
% \begin{center}
% \fbox{\rule{0pt}{2in} \rule{.9\linewidth}{0pt}}
\includegraphics[width=1.0\textwidth]{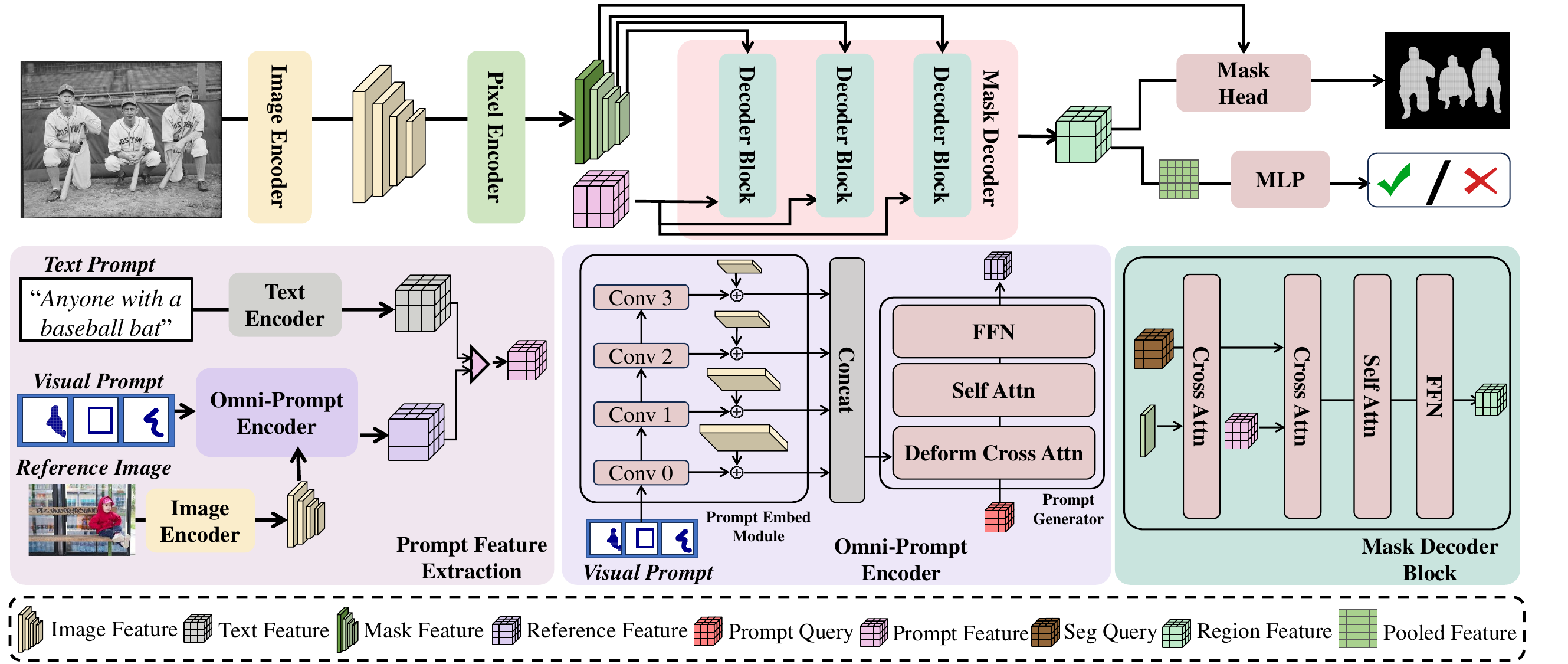}
% \vspace{-1em}
\caption{The architecture of the proposed baseline model, termed  OmniSegNet. The target image is processed by the image encoder and the pixel encoder to obtain multi-scale visual features for accurate mask decoding. When input text prompts only, the text features extracted by the text encoder will be directly fed to the mask decoder for cross-modal interaction. OmniSegNet also introduces a novel omni-prompt encoder to handle the spatial prompts, such as masks, boxes and scribbles. With this omni-prompt encoder, the semantics of the visual referent in the image can be extracted to guide mask decoding. OmniSegNet can also support the processing of visual and text prompts at the same time.}
\label{a_framework}
% \vspace{-1em}
% \end{center}
\end{figure*}
In this appendix, we first provide additional details on the OmniSegNet method, including the Omni-Prompt Encoder for encoding visual prompts and the Mask Decoder for decoding the target mask. We then describe the model's implementation, including the encoder architecture, parameter configuration, and training loss. 
% In ablation experiments, we analyze the impact of different visual prompt types, \emph{Prompt Embed Module} (PEM) module fusion methods, and training sample ratios on model performance, validating the importance of each component. Finally, we evaluate the generalization capabilities of OmniSegNet on the existing RIS benchmark dataset. Experimental results show that it outperforms existing methods on multiple datasets, demonstrating excellent multimodal understanding and segmentation capabilities.
For ablation experiments, the analyses of visual prompt types and \emph{Prompt Embedding Module} (PEM) fusion strategies are presented in the main paper. This supplement reports the remaining ablation, focusing on the impact of the sampling ratio between text- and visual-guided data during training, further validating the significance of this design choice.

\section{Method}
\subsection{Omni-Prompt Encoder}
To handle the effective fusion of visual prompts, we propose a novel \emph{omni-prompt encoder} for OmniSegNet, which consists of two modules, namely \emph{Prompt Embed Module} (PEM) and \emph{Prompt Generator}.
\\
\textbf{Prompt Embed Module (PEM).}
As shown in \cref{a_framework}, the visual prompt is initially processed by the PEM to generate the visual prompt encoding. The PEM employs its internal convolution layer to map the input visual prompt into the same semantic space as the reference multi-scale visual features \( \mathbf{F}_{r} \), achieving feature alignment. 
% In each convolution layer \( \mathbf{Conv}_{i} \), the encoded visual prompt \( \mathbf{P}^{(i)} \) is element-wise added to the reference visual features at the corresponding scale, where \( i \in \{0,1,2,3\} \). 
In each convolution layer \( \mathbf{Conv}_{i} \), the visual prompt \( \mathbf{P}^{(i)} \) is first passed through a convolutional transformation, and the resulting features are then element-wise added to the reference visual features at the corresponding scale, where \( i \in \{0,1,2,3\} \).
This operation can be expressed as:
\begin{equation}
\mathbf{F}'^{(i)}_{s} = \mathbf{F}_{r}^{i}+\mathbf{Conv}_{\mathrm{i}}(\mathbf{P}^{(i)}),
\end{equation}  
Here, \( \mathbf{F}'^{(i)}_{s} \) represents the reference visual features at scale \( i \). To further integrate multi-scale information, the set of scaled features \( \mathbf{F}'^{(i)}_{s} \) across different scales is flattened and concatenated along the channel dimension. This process yields a unified reference feature representation \( \mathbf{F}'_{r} \) that captures rich multi-scale context for subsequent processing stages.
\\
\textbf{Prompt Generator.}
The Prompt Generator extracts discriminative prompt features via a compact three-layer architecture, as illustrated in Fig.\ref{a_framework}. Each layer consists of \emph{deformable cross-attention}~\cite{kamath2021mdetr}, \emph{self-attention} and a \emph{feed-forward network} (FFN). Given a set of learnable prompt queries \( \mathbf{F}_{q} \in \mathbb{R}^{n \times c} \), where \( n \) denotes the number of queries and \( c \) is the feature dimension, the module first performs cross-attention with the reference features \( \mathbf{F}_{r}^{'} \) from the PEM to capture essential object semantics:
\begin{equation}\label{eq3}
\mathbf{F}_{q}^{'} = \operatorname{DeformCrossAttn}(\mathbf{F}_{q}, \mathbf{F}_{r}^{'}),
\end{equation}
The resulting query features \( \mathbf{F}_{q}^{'} \) are then refined through self-attention to model global dependencies and passed through an FFN for transformation:
\begin{equation}\label{eq4}
\mathbf{F}_{p} = \operatorname{FFN}(\operatorname{SelfAttn}(\mathbf{F}_{q}^{'})).
\end{equation}
After three stacked layers of such processing, the final visual prompt features \( \mathbf{F}_{p} \) are obtained, which effectively encode semantic prompts from the reference and guide foreground segmentation in the target image.
\subsection{Mask Decoder}
The Mask Decoder consists of three sequential blocks, each comprising two cross-attention layers, one self-attention layer and one FFN. The decoder takes as input: (1) a learnable segmentation query $\mathbf{F}_{seg}$, (2) a prompt feature $\mathbf{F}_{pt}$ (either from the text or visual reference), and (3) multi-scale mask features $\mathbf{F}^i_m$ from the target image, where $i \in \{1,2,3\}$. In each block $i$, the process is as follows:
\begin{align}
\mathbf{F}_{seg} &= \text{CrossAttn}(\mathbf{F}_{seg}, \mathbf{F}^i_m), \\
\mathbf{F}_{seg} &= \text{CrossAttn}(\mathbf{F}_{seg}, \mathbf{F}_{pt}), \\
\mathbf{F}_{seg} &= \text{FFN}(\text{SelfAttn}(\mathbf{F}_{seg})).
\end{align}
The updated $\mathbf{F}_{seg}$ is propagated to the next block. After the third block, the final $\mathbf{F}_{seg}$ is treated as the region feature $\mathbf{F}_{reg}$. This feature is fused with the low-level visual feature $\mathbf{F}^0_m$ and passed to a MaskHead to produce the final segmentation mask. Additionally, $\mathbf{F}_{reg}$ is processed by an MLP to generate the no-target prediction.

\subsection{Loss Setting}
Our framework employs three training objectives for supervised optimization to enhance the segmentation performance. Drawing on the approach in~\cite{liu2023gres}, we introduce the cross-entropy loss as the basic loss function. Firstly, we define \( \mathcal{L}_{mask} \) to measure the difference between the ground truth label \( \mathbf{M}_{gt} \) and the final predicted mask \( \mathbf{M}_{pre} \), ensuring that the model accurately identifies the target region.
Secondly, we design \( \mathcal{L}_{region} \) to optimize the learning of the regional feature \( \mathbf{F}_{reg} \). This loss measures the degree of matching between the regional feature \( \mathbf{F}_{reg} \) and the corresponding regional mask \( \mathbf{M}_{region} \). Here, \( \mathbf{M}_{region} \) is adjusted to the same scale as \( \mathbf{F}_{reg} \) by performing linear interpolation on the ground truth mask \( \mathbf{M}_{gt} \). 
Additionally, to further improve the model's ability to detect the presence of the target, we apply global average pooling to all regional features \( \mathbf{F}_{reg} \) and predict the target’s existence \( \mathbf{P}_{gt} \) based on the pooling results \( \mathbf{P}_{pre} \). To this end, we introduce the \( \mathcal{L}_{nt} \) loss function, which optimizes the accuracy of the target existence prediction. Therefore, the total loss with balance factor $\lambda$ is:
\begin{equation}\label{eq5}
  \mathcal{L}_{total} = \lambda_{1}\mathcal{L}_{msak} + \lambda_{2}\mathcal{L}_{region} +\lambda_{3}\mathcal{L}_{nt}.
\end{equation}

\section{More Experiments} 
\subsection{Implementation Settings} 
We adopt the Omni-Prompt Encoder as our core architecture. We employ three sets of deformable cross-attention, self-attention and FFN modules within its Prompt Generator, each with a feature dimension of $256$. To ensure alignment between text and visual prompt features, the length of Prompt Query is set to the predefined maximum sentence length, enabling both modalities to share the same Mask Decoder. The Mask Decoder is composed of nine mask decoder blocks, grouped into three groups, each corresponding to one scale of features extracted by the Pixel Encoder. 
We use Swin-B~\cite{liu2021swin} as the image encoder and BERT~\cite{devlin2018bert} as the text encoder with a hidden dimension of $768$. During training, the input images are resized to $480 \times 480$ and the maximum sentence length is fixed at $20$. The model is optimized using AdamW~\cite{loshchilov2017decoupled} with a weight decay of $0.05$. The initial learning rate is set to $1e-5$ and follows a polynomial decay schedule.

\begin{table}[t]
\centering
\caption{Ablation study of the batch sample ratios in our training regimes.}

\resizebox{\columnwidth}{!}{%

\begin{tabular}{c|ccc|ccc}
\toprule
\multirow{2}{*}{\makecell{\centering Ratio}} & \multicolumn{3}{c|}{gRefCOCO-val} & \multicolumn{3}{c}{Visual-test}    \\
\cmidrule(lr){2-4} \cmidrule(lr){5-7} 
                & cIoU & gIoU & N$_{acc}$ &  cIoU & gIoU & N$_{acc}$   \\
\midrule
            \emph{-} & 64.21 & 65.53 & 58.72 & 75.35 & 66.36 & 89.48  \\
            \emph{5/3} & 63.98 & 65.36 & 59.37 & 76.34 & 67.61 & 90.12  \\
% \midrule
           \emph{9/4} & 64.50 & 66.00 & 59.78 & 76.29 & 67.79  & 89.86 \\
% \midrule
            \emph{7/3} & 64.22 & 65.61 & 57.64 & 76.39 & 68.12 & 89.77  \\
    
            \emph{5/2} & 64.10 & 66.49  & 61.48 & 76.51 & 68.30 & 90.23 \\
            \emph{3/1} & 64.75 & 66.81  & 61.43 & 76.51 & 68.22 & 89.91 \\
            \emph{7/2} & \textbf{65.30} & \textbf{67.57} & \textbf{63.44} & \textbf{76.63} & \textbf{68.87} & \textbf{90.81}  \\

\bottomrule
\end{tabular}%

}

\label{ratio}
\end{table}
\subsection{Ablation Study of OmniSegNet}
\textbf{Effect of Batch Sample Ratio.}
\cref{ratio} presents a comparative analysis of using different batch sample ratios during joint training of the gRefCOCO dataset and visual prompt data from the OmniRef benchmark. The evaluation datasets are their respective test sets, and the impact of varying ratios on gRefCOCO-val and Visual-test performance is assessed.
Notably, the first row (``–'') reports the results of training the model independently on each dataset—gRefCOCO for text-guided referring segmentation, and OmniRef visual prompt data for visual-guided segmentation. These results provide a baseline for understanding how much joint training contributes beyond single-modality supervision.
% We observe that standalone training achieves reasonable performance for each modality, but the two types of supervision do not benefit from each other in this isolated setting.
%
For gRefCOCO-val, increasing the proportion of text-guided samples significantly improves performance, indicating that rich language supervision helps the model better understand and express natural language referring cues. Compared with the ``–'' baseline, nearly all mixed ratios produce higher cIoU, gIoU, and N$_{acc}$, demonstrating that incorporating visual-guided samples does not harm text-guided performance and can even improve generalization.
On Visual-test, adding more visual-guided samples improves performance, but the gains remain relatively stable across different ratios. However, joint training consistently outperforms the standalone (``–'') baseline, suggesting that text-guided signals also provide auxiliary benefits for visual prompt understanding.
In summary, the \emph{7/2} ratio achieves the best performance in both tasks, balancing text and visual prompt learning. These observations demonstrate that joint training is not a simple combination of two tasks; instead, it enables beneficial cross-modality transfer, effectively surpassing the standalone (``–'') baselines and confirming the value of strategic data distribution in multimodal segmentation training.

\end{document}